\documentclass{article}
\usepackage{jmlr2e}
\editor{}

\firstpageno{1} 

\usepackage[utf8]{inputenc} 
\usepackage[T1]{fontenc}    
\usepackage{hyperref}       
\usepackage{url}            
\usepackage{booktabs}       
\usepackage{amsfonts,amssymb}       
\usepackage{nicefrac}       
\usepackage{microtype}      
\usepackage{xcolor}         
\usepackage{amsmath}        
\usepackage{longtable}      
\usepackage{siunitx} 
\usepackage{graphicx}

\title{BanditCAT and AutoIRT: Machine Learning Approaches to Computerized Adaptive Testing and Item Calibration}

\author{
    \name James Sharpnack \email james.sharpnack@duolingo.com\\
    \name Kevin Hao \email kevin.hao@duolingo.com\\
    \name Phoebe Mulcaire \email phoebe.mulcaire@duolingo.com\\
    \name Klinton Bicknell \email klinton.bicknell@duolingo.com\\
    \name Geoff LaFlair \email geoff.laflair@duolingo.com\\
    \name Kevin Yancey \email kevin.yancey@duolingo.com\\
    \name Alina A. von Davier \email alina.vondavier@duolingo.com\\
    \addr Duolingo, 5900 Penn Ave., Pittsburgh, PA 15206
}

\newcommand{\cl}{\mathcal}
\newcommand{\bb}{\mathbb}

\newcommand{\bo}{\mathbf}

\begin{document}

\maketitle

\begin{abstract}
In this paper, we present a complete framework for quickly calibrating and administering a robust large-scale computerized adaptive test (CAT) with a small number of responses.
Calibration --- learning item parameters in a test --- is done using \emph{AutoIRT}, a new method that uses automated machine learning (AutoML) in combination with item response theory (IRT), originally proposed in \cite{sharpnack2024autoirt}.
AutoIRT trains a non-parametric AutoML grading model using item features, followed by an item-specific parametric model, which results in an explanatory IRT model.
In our work, we use tabular AutoML tools (\texttt{AutoGluon.tabular}, \cite{erickson2020autogluon}) along with BERT embeddings and linguistically motivated NLP features.
In this framework, we use Bayesian updating to obtain test taker ability posterior distributions for administration and scoring.

For administration of our adaptive test, we propose the \emph{BanditCAT} framework, a methodology motivated by casting the problem in the contextual bandit framework and utilizing item response theory (IRT). 
The key insight lies in defining the bandit reward as the Fisher information for the selected item, given the latent test taker ability ($\theta$) from IRT assumptions.  
We use Thompson sampling to balance between exploring items with different psychometric characteristics and selecting highly discriminative items that give more precise information about $\theta$.  
To control item exposure, we inject noise through an additional randomization step before computing the Fisher information.
This framework was used to initially launch two new item types on the DET practice test using limited training data.
We outline some reliability and exposure metrics for the 5 practice test experiments that utilized this framework.
\end{abstract}

\section{Introduction}
 
The goal of test assessment is to measure an abstract characteristic, like listening comprehension or speech production, and summarize the test taker's proficiency with a score. 
Computerized adaptive tests sequentially administer items (i.e., questions) according to an administration policy.
Ideally, this policy maximizes the information gathered about the test taker's ability, e.g., by adjusting the difficulty of administered items based on previous test taker responses. 
Reported scores are designed to reflect the test taker's proficiency, which is achieved through proper calibration of the test items.
In this paper, we introduce two methods for calibrating and deploying adaptive tests.

{\bf Calibration.}
Item response theory (IRT) models an item’s psychometric characteristics, also known as item parameters, such as difficulty and discrimination, which are essential for adaptive testing and scoring.
Item Response Theory (IRT) focuses on the measurement and evaluation of educational or psychological latent traits, that is, unobservable variables, such as ability, skill, or competence levels \citep{lord1950statistical}.
However, traditional IRT calibration requires extensive data collection, typically gathered during an item piloting phase. 
There are three primary situations in which we might calibrate a proportion of an item bank, cold-start and jump-start piloting of new items, and warm-start recalibration of the entire item bank.

\begin{enumerate} 
\item \emph{Cold-start:} New pilot items are added without response data, and only item features are used to calibrate them, while re-calibrating the existing operational item bank. 
\item \emph{Jump-start:} A limited number of responses are collected for new pilot items in a piloting phase, while both pilot and operational items are calibrated. 
\item \emph{Warm-start:} The operational item bank is re-calibrated to reflect recent response data, often due to changes in the user interface, test taker population, or preparation materials. 
\end{enumerate}
When an entirely new item bank is to be launched, there is no existing data with which to calibrate the items, and typically we have a pre-launch phase where we gather responses with which to calibrate items.
Explanatory IRT models refer to item response theory (IRT) models that use item features to predict item parameters, such as difficulty or discrimination, rather than treating each item independently.
This enables calibrating tests in the cold and jump-start settings, by using item level features (such as NLP features) to predict the item parameters.

In this work, we outline AutoIRT, a new approach to calibrating item parameters using automated machine learning (AutoML). 
AutoML tools train machine learning models with automated tuning parameter selection, feature engineering, data processing, and task selection.
AutoIRT leverages AutoML to train IRT models from test responses and item content, overcoming the need for manual tuning and supporting multi-modal input. 
While traditional AutoML models are not inherently interpretable, AutoIRT maintains compatibility with IRT models, the standard in psychometrics. 
AutoIRT was introduced in \cite{sharpnack2024autoirt} and is the first use of AutoML to fit IRT models.

{\bf Adaptive Testing.}
A computerized adaptive test (CAT) is a psychological or educational test that is usually administered in a digital environment and it uses an iterative, algorithm-based approach to select and administer test items. 
In IRT-based CATs, items are chosen based on the TT's estimated ability level throughout the testing process, and the estimated ability is continuously updated after each item is responded to. 
Test developers aim to devise optimal sequences of items for each TT, so that the accuracy of the test is maximized, while keeping the test relatively short, the item exposure low, and the cheating rates by copying from other TT low. 
Hence, a CAT is a much more efficient test, being shorter while maintaining the accuracy of linear tests \citep{lord1980, wainer2000cat}. 
Although a CAT can be administered outside of an IRT framework, it is customary in psychometrics to use IRT for every part of the testing process: item bank calibration, item routing, and ability estimation. 

In this work, we introduce a new approach to item selection in a CAT, called BanditCAT. 
This framework casts CAT in the contextual bandit framework where the rewards are the Fisher information of that item for the test taker's ability.
The Fisher information is derived from the IRT item parameters that were estimated using AutoIRT.
We use Thompson sampling along with an additional randomization step to control for exposure while still selecting informative items.
We show the effectiveness of this system by outlining its use on the practice Duolingo English Test during the launch of two new item types, Y/N Vocabulary and Vocabulary-in-context.
This preliminary report introduces the first method in the BanditCAT framework, but it does not exploit the full capability of this approach.
A more complete follow-up work based on this framework is forthcoming.

\subsection{Background on Computerized Adaptive Testing}

{\bf Item Response Theory.}
IRT corresponds to a class of interpretable statistical models that is commonly used for the scoring and administration of psychological and educational assessments, including computerized adaptive tests (CATs) \citep{baker2004item, lord1980, wainer2000cat}.
While there are multidimensional IRT models, in most operational applications the unidimensional IRT models are typically used.
The unidimensional IRT framework makes the following assumptions: the test taker (TT) grade distribution depends on a single parameter $\theta$ (unidimensionality), the expected grade is increasing in ability (monotonicity), and grade measurements are independent given that the TT was administered the items (local independence) \citep{bock1981mml}.
IRT models the grades and ability $\theta$ for a given TT.
Define $G_i$ as the random grade if they were administered item $i$, then the item response function is the expectation $p_i(\theta) = \bb E[G_i | \theta]$ (by unidimensionality).
A test session consists of a sequence of $T$ administered items $\bo I_T = (I_1, I_2, \ldots, I_T)$ and observed grades $\bo G_T = (G_1, \ldots, G_T)$.
The local independence assumption implies that these are both random with the following Markov structure: $G_t$ is independent of $\bo G_{t-1}$ given $I_t$ for $t = 1, \ldots, T$ \citep{baker2004item}.

One important aspect of IRT models is that they are \emph{interpretable}, which is achieved by specifying a model.
The most commonly used IRT models are the parametric logistic models for binary grades (right or wrong response).
In these models the grade probability, also known as the item response function (IRF), takes the form,
\begin{equation}
\label{eq:3PL}
    p(\theta; \phi_i) = \bb P \{ G_t = 1 | \theta, I_t = i \} 
    = c_i + (1-c_i) \cdot \sigma (a_i (\theta - d_i)),
\end{equation}
where $\phi_i = (a_i, c_i, d_i)$ are the item parameters: slope, or discrimination, $a_i$, chance parameter, $c_i$, and difficulty, $d_i$ \citep{lord1980, fischer1973lltm}.
The slope parameter is related to the efficiency of an item in terms of the information provided to distinguish among test takers.
The chance parameter is part of the 3-parameter logistic IRT model, reflecting the probability of a correct response due to chance, and the difficulty parameter reflects the difficulty of the item for a particular TT population.
There is a clear connection between the IRF and the item parameters.
For example, at larger discrimination levels, the IRF is steeper, and a small increase in ability around that item's difficulty level will lead to a significant increase in the probability of a correct response \citep{baker2004item, fox2010bayesian}.

{\bf Calibration with explanatory IRT Models.}
Without additional information about items, calibrating IRT models such as \eqref{eq:3PL} requires hundreds of responses.
In order to calibrate in the cold-start and jump-start settings, we need to use contextual information about the items, typically derived from the content of the item itself (e.g., lexical and semantic content).
We assume throughout that each item has associated explanatory features, denoted as $\bo x_i \in \bb R^d$. 
These features may be derived from neural network embeddings such as BERT \citep{devlin2019bert} or CLIP \citep{radford2021learning}, hand-crafted linguistic features, or other key characteristics of the item. 
The use of item features in explanatory IRT models has a long history. 
A prominent example is the Log-Linear Trait Model (LLTM) by \cite{fischer1973lltm}, which extends the Rasch model by using item features to predict item difficulty through a linear model (see also \cite{deboeck2004frameworks}). 
More recent extensions leverage pretrained language models like BERT \citep{devlin2019bert} to calibrate items from limited response data. 
For instance, \cite{benedetto2021transformers} predicted item difficulty using a fine-tuned BERT model based on student responses to multiple-choice questions, 
while \cite{byrd2022predicting} employed a similar approach to estimate both difficulty and discrimination parameters for 2PL IRT models from natural language questions. 
Their work incorporated linguistically motivated features such as semantic ambiguity, alongside contextual embeddings generated by BERT. 
Additionally, \cite{reyes2023multiple} used RNNs to predict item difficulty, which had been pre-fitted using the Rasch model. 
Most of these approaches focus on predicting item parameters that have been pre-estimated using a large item bank with substantial response data. 
Training such models requires an extensive item bank, with hundreds of responses per item to generate the necessary training data.

An alternative method is to train an explanatory IRT model by fitting the item parameters through a neural network (NNet), where the final layer represents the IRT model \eqref{eq:3PL}.
This approach trains a NNet to predict scores from features and ability $\theta$, while constraining the model to follow the IRT structure.  
BERT-LLTM was introduced in \cite{mccarthy2021jump} to train LLTM models using BERT embeddings alongside hand-crafted linguistic features.  
More recently, \cite{yancey2024bert} proposed BERT-IRT, which fits a 2PL model by constraining the NNet to follow the form of \eqref{eq:3PL} with $c_i = 0$ for all items $i$.  
In that work, they demonstrate that BERT-IRT can accurately train explanatory IRT models with only a few responses per item and apply it to English language proficiency assessments.  
They use a linear form for $\log(a_i)$ and $d_i$, which \cite{sharpnack2024autoirt} enhanced by employing a fully non-parametric AutoML model.  
The model is trained using $\bo x_i$ and $\tilde \theta_s$ as inputs, with the predicted binary score as the output.  
In our experiments we used a use a proxy estimate for test taker ability, $\tilde \theta_s$, which is derived from scores on other item types (similar to the approach in \cite{yancey2024bert}).

{\bf Computerized Adaptive Testing.} 
There are several broad categories of CAT algorithms, such as maximum information selection \citep{birnbaum1968some}, maximum global information selection \citep{chang1996global}, fully Bayesian selection \citep{vanderlinden1998bayesian}, and shadow testing \citep{vanderlinden2005hybrid}.
Most of these methods base their selection on the Fisher information of each item given our current estimate of $\theta$ for the model \eqref{eq:3PL}.
For the 3PL model this is,

\begin{equation}
\label{eq:fisher_3pl}
F_i(\theta) = \frac{\left(a_i \cdot (1 - c_i) \cdot p_2(\theta; a_i, d_i) \cdot (1 - p_2(\theta; a_i, d_i))\right)^2}{\left(c_i + (1 - c_i) \cdot p_2(\theta; a_i, d_i)\right) \cdot \left(1 - \left(c_i + (1 - c_i) \cdot p_2(\theta; a_i, d_i)\right)\right)}
\end{equation}

where
\begin{equation}
p_2(\theta; a_i, d_i) = \frac{1}{1 + \exp(-a_i(\theta - d_i))},    
\end{equation}
is the 2PL IRF.
The main exception to this are those that select based on the KL divergence \citep{chang1996global}.
All of these methods require provisional estimates of $\theta$, furthermore several require estimating the posterior distribution of $\theta$ given the responses observed thus far for the given test session.
Luckily, because the IRT models are univariate, the posterior can be estimated without trouble by approximating distributions as point masses on a discrete grid of $\theta$ values.
In multi-dimensional models, this approach breaks down and more sophisticated Bayesian updating methods are needed \citep{mulder2010multidimensional}.

Directly selecting items that maximize this information measure can result in poor exposure rates, namely, a handful of items can be selected far too frequently.
Several methods have been proposed for exposure control including item cloning \citep{vanderlinden2000adaptive} and the Sympson-Hetter method \citep{sympson1985controlling}.
While the latter tracks the current exposure rates and guides them toward a target, the former uses randomization to balance exposure by randomly selecting items from the top k most informative.
The randomization approach has a practical advantage in production systems because they are not dependent on any global statistics that need to be updated, and so can be deployed to servers independently who do not need a common cache.


{\bf Thompson Sampling and Contextual Bandits.}
A stochastic contextual bandit problem models a repeated interaction between a player and an environment. 
In each round, the player is presented with information about $K$ actions (called arms), typically represented by $d$-dimensional feature vectors. 
The player must select an arm based on past observations. 
Only the reward from the selected arm is revealed, and the relationship between the rewards and features is governed by a linear model. 
The objective of the player is to maximize cumulative reward over $T$ rounds. 
Over the past few decades, bandit algorithms have gained widespread use in various real-world applications, such as recommender systems \citep{li2010contextual}, online advertising \citep{schwartz2017customer}, and clinical trials \citep{woodroofe1979one}.

Classic stochastic linear contextual bandit algorithms, such as Linear Upper Confidence Bound (LinUCB) \citep{li2010contextual} and Linear Thompson Sampling (LinTS) \citep{agrawal2013thompson}, have been shown to achieve nearly optimal total reward, given some modeling assumptions.
Like most bandit algorithms, the objective is to begin by exploring the space of actions to get a sense of which are profitable, and then exploiting the actions that will give the highest reward.
Thompson sampling accomplishes this by modeling reward as a function of the context vectors and latent parameters.
By tracking the posterior distribution of the latent parameters, it draws realizations of these from the posterior and then selecting actions that maximize expected reward evaluated at that draw.
Translating this to the CAT setting, the `player' is the CAT, `environment' is the test taker, `actions' are the items administered, `context' is based on the previous responses of the test taker and information about the items themselves, `reward' is the information gained about test taker ability.

\section{Item Calibration with AutoIRT}

To harness the flexibility and performance of AutoML, we begin by fitting a grade classifier using the ability parameter $\bo \theta$ and $d$-dimensional item features $\bo x_i \in \bb R^d$ as inputs. 
In our experiments, we employ a stacked ensemble of models—random forests, LightGBM, XGBoost, and CATBoost—implemented through the AutoGluon-tabular Python package \cite{erickson2020autogluon}. 
This package was chosen for its strong performance in tabular data benchmarks \cite{gijsbers2024amlb}. 
Although we focus on tabular models since we have already engineered features like BERT embeddings, AutoML supports multimodal input. 
We hope to eventually apply AutoIRT directly to raw item content. 
The item ID is passed to the AutoML predictor as a feature, similar to the use of random effects in traditional models. 

Let $\hat p(\theta; \bo x_i)$ represent the AutoML-predicted probability of a correct response for a test taker with ability parameter $\theta$ and an item with features $\bo x_i$. 
Next, we extract a more interpretable IRT model by projecting the AutoML model onto the closest IRT model using a least squares approach. 
Specifically, we minimize the following loss function to estimate item parameters:
\[
    L(\bo \phi) = \sum_{i \in \cl R} \sum_{\theta \in \Theta} (p(\theta; \phi_i) - \hat p(\theta; \bo x_i))^2,
\]
where $\Theta$ is a regular grid of $\theta$ values. 
This ensures that $\Theta$ covers most of the probability mass of the distribution of $\theta$s. 
In the case of the 3PL model \eqref{eq:3PL}, we either will allow the chance parameter $c_i$ to vary, or be fixed (typically at $c_i = 0$, the 2PL model).
In \cite{sharpnack2024autoirt}, an extension of this model was proposed which uses the above method as the M-step in a Monte Carlo EM algorithm.
This allowed them to jointly learn $\theta$ and the item parameters $\phi_i$.
One restriction is that the chance parameters and $\theta$ cannot be jointly learned due to a lack of identifiability \citep{baker2004item}, so that approach required fixing chance $c_i$.
In our setting, we initialize $\theta$ to be a weighted combination of other item type scores, resulting in a proxy ability.
This approach is similar to what was taken in \cite{yancey2024bert}.

\section{BanditCAT Framework}

\subsection{Contextual Bandit Interpretation of CATs}

In this section, we will cast CAT administration as a contextual bandit problem.
Each arm in the bandit setting corresponds to an item $I_t$ and the action is the administration of the item at round $t$ of $T$ rounds.
Each item has context, which are the item features, $x_i \in \bb R^d$.
The reward is the Fisher information for that item $F_{I_t}(\theta)$ for that test taker's true $\theta$.
This reward is unobserved because we do not have access to the true $\theta$, but we are able to model the reward via our provisional estimate of $\theta$ (or the current posterior) and our calibrated item parameters.
The grades enter into the picture via our provisional estimate of $\theta$, and our understanding of how informative an item is for a test taker's ability is a direct product of the observed grades, $G_{I_t}$.

One of the advantages of using the Fisher information is that it directly measures the information content of a response to that item.
At the end of a test session we can evaluate how effective our item selections were by measuring the information content of the data that we gathered about the test taker.
Specifically, the total reward for that session is the Fisher information of all data gathered thus far,
\begin{equation}
\label{eq:total_fisher}
    \bar F_T(\theta) := \bb E \left[ \left( \sum_{t=1}^T \frac{\partial}{\partial \theta} \log f(G_{I_t} | \theta) \right)^2 \right] = \sum_{t=1}^T F_{I_t}(\theta).
\end{equation}
The above follows from the fact that the score (gradient of log likelihood) at the true $\theta$ has mean 0 and the local independence assumption.
It should be noted that we are conditioning on the items that were administered, $\bo I_n$, in that session.
We know by the Cramer-Rao lower bound that the variance of any unbiased estimator for $\theta$ is at least $1 / F_T(\theta)$, under regularity conditions, that are satisfied in this case.


For the 2PL model, there is a clear relationship between the item parameters and the Fisher information, namely the height is $a_i^2 / 4$, regardless of $d_i$.
This is not true of the 3PL model, since the chance parameter makes the maximum information point deviate from $d_i$.
The height is then a complex function of the item parameters, which will not be amenable to our exposure control randomization method.
To this end, we approximate the 3PL Fisher information with a Gaussian kernel,
\begin{equation}
\label{eq:info_kernel}
    F_i(\theta; \mu_i, \nu_i, h_i) = h_i \cdot \exp \left( - \frac{(\theta - \mu_i)^2}{2 \nu_i^2} \right),
\end{equation}
where $\mu_i$ is the center parameter, $h_i$ is the height, and $\nu_i$ is the bandwidth parameter.
For a given set of item parameters $a_i, c_i, d_i$ we obtain $\mu_i, h_i, \nu_i$ by moment matching.
For the 2PL model, we use \eqref{eq:fisher_3pl} directly, which simplifies greatly.

\subsection{Thompson Sampling with Exposure Control}

In this section, we introduce an initial algorithm for item selection under the BanditCAT framework, which we call BanditCAT V1 to indicate that it is the first incarnation of this approach.
In our setting the reward $F_{I_t}(\theta)$ is unobserved, however we can build a model for it via the posterior $\pi(\theta | G_{\bf I_t})$.
To utilize the bandit framework, we let the reward model be the following, 
\begin{equation}
    \hat r_i(\theta) = F_i(\theta ; \hat \phi_i),
\end{equation}
where $\hat \phi_i$ is the predicted item parameters $(\hat a_i, \hat c_i, \hat d_i)$ from AutoIRT.
In the case of the Gaussian kernel Fisher information \eqref{eq:info_kernel} we first convert the item parameters $\hat \phi_i$ into kernel parameters $(\hat h_i, \hat \mu_i, \hat \nu_i)$.
In order to utilize the full contextual bandit framework, the item parameters $\hat \phi_i$ should be drawn from a distribution, which we reserve for a future work.

We want to control the frequency of arm selection (i.e., exposure) for Thompson sampling in bandits, and we accomplish this using randomization in the item parameters when calculating $\hat r_i(\theta)$.
Specifically, for the 2PL model, we sample the discrimination parameter,
\begin{equation}
    \label{eq:2PL_randomization}
    \tilde a_i \sim {\rm Gamma} (\hat a_i / \gamma, \gamma),
\end{equation}
where $\gamma > 0$ is a global scale parameter (the mean for the Gamma is $\hat a_i$).
For the 3PL model we use the Fisher information approximation, \eqref{eq:info_kernel}, and instead add randomness to the height parameter,
\begin{equation}
    \label{eq:3PL_randomization}
    \tilde h_i \sim {\rm Gamma} (\hat h_i / \gamma, \gamma).
\end{equation}
These randomization approaches were developed through additional experimentation and simulation.
In our method, we draw $\theta$ from the posterior distribution given the observed grades.
However, we additionally allow for multiple $\theta_k$'s to be drawn and then we average the estimated rewards for each of these.
The purpose of this is to interpolate between traditional Thompson sampling and fully Bayesian item selection \citep{vanderlinden1998bayesian}.
Specifically, the expectation of $F_i(\theta)$ over the posterior is the Bayesian information criteria, so this is approximated by the average with enough Monte Carlo samples of $\theta$.

\subsection{BanditCAT V1}

{\bf Input:} eligible item bank $\cl I$, fitted item parameters $\hat \phi_i$ (from AutoIRT), prior distribution over $\theta$, $\pi_1$, and exposure control parameter $\gamma > 0$.  

At each round, $t = 1,\ldots,T$,
\begin{enumerate}
    \item For $k=1,\ldots,K$, independently draw $\theta_k \in \Theta$ from $\pi_t(.)$.
    \item For each test item, $i \in \cl I$, draw information parameters,
    \begin{enumerate}
        \item For the 3PL model, draw $(\tilde h_i, \hat \mu_i, \hat \nu_i)$ from \eqref{eq:3PL_randomization} to obtain $\tilde r_i(\theta_k), k=1,\ldots,K$, according to \eqref{eq:info_kernel}.
        \item For the 2PL model, draw $(\tilde a_i, \hat d_i)$ from \eqref{eq:2PL_randomization} to obtain $\tilde r_i(\theta_k), k=1,\ldots,K$, according to \eqref{eq:fisher_3pl}.
    \end{enumerate}
    \item Select item $I_t \in \cl I$ that maximizes $\frac{1}{K} \sum_{k=1}^K \tilde r_i(\theta_k)$.
    \item Observe the grade $G_{I_t}$ for that test and update posterior $\pi_t(.) = \pi(.|G_{I_1},\ldots,G_{I_t})$.
\end{enumerate}
Finally, we return the posterior mean $\bb E_{\pi_T}[\theta]$ as the score for that item type.

Aside from its statistical soundness there are some advantages to this method.
Particularly, it can be deployed independently to any server without any shared cache or memory, making server autoscaling quite simple.
Computationally the method is quite simple, with the most expensive operation being the very fast posterior calculation.
In the instance that there are additional eligibility criteria, then BanditCAT V1 will reduce the eligible item bank prior to step (2).
For multiple item types, this method is applied in sequence for each item type.
Multidimensional variants of BanditCAT are in development as well as a complete Thompson sampling approach that samples the item parameters as well.

\subsection{CAT Simulation}

For model selection and tuning config parameters (most notably the exposure control parameter $\gamma$), we need to be able to simulate what will happen when we apply our administration algorithm to new test takers.
We use the following \textbf{nearest neighbor matching algorithm} to simulate a new session for single text vocab.

{\bf Input:} Test sessions with proxy $\theta$'s derived as a weighted average of scores from item types other than those being modified and responses for the item types in question.

For a sample of real tests, calculate their proxy $\theta$s (call this $S_0$) and for each test do the following:
\begin{enumerate}
    \item Administer the first item, $I_1$, for a session.
    \item For each round $t = 1, \ldots, T$,
    \begin{enumerate}
        \item Find the historical session that was administered item $I_t$ that has the closest proxy $\theta$ to the target $S_0$ (nearest neighbor matching).
        \item Let the grade $G_t$ for the simulated session be the grade for the matched historical session on item $I_t$.
        \item Update posterior and select next item $I_{t+1}$.
    \end{enumerate}
    \item Score this simulated session as if it were real
\end{enumerate}
We can use these simulated sessions to evaluate our desired metrics, which is typically the exposure rates so that we can hit our target.

\subsection{Assessing Vocabulary in the Duolingo English Test}

In this work, we study the calibration and administration of vocabulary item types in the Duolingo English Test (DET).
We focus on two item types that are at the start of the DET which assess vocabulary knowledge.
These items in combination assess the form, meaning, and use of English words.

\textbf{Y/N Vocabulary and Vocab-in-Context item types.}
We focus on the first two item types in the DET (V8): yes/no vocabulary (Y/N Vocab) and vocabulary in context (ViC).
The practice test includes 3290 Y/N Vocab items and 585 ViC items, with 18 Y/N Vocab and 9 ViC items administered per test session in succession.
Test takers have 5 seconds to respond to Y/N Vocab items and 20 seconds for ViC items.
Y/N Vocab requires the test taker to identify whether a word is real or fake (with fake words generated by an RNN).
ViC is a fill-in-the-blank task where the test taker completes a word within a sentence.
For example, a Y/N Vocab item might ask if ``newbacal'' is a real word, while a ViC item might ask for the missing word in “I’m sorry for the inter\_\_\_\_\_\_\_, but could you explain that last part again?”
Additional details are provided in \cite{sharpnack2024autoirt}, and practice tests can be taken for free at \texttt{http://englishtest.duolingo.com}.

{\bf Item Features.} Throughout we use a combination of linguistic and features from a foundation model (RoBERTa).  For each item, we calculate a set of features based on its content, using the Corpus of Contemporary American English (COCA), \cite{davies2008word}, for corpus-based features. In Y/N Vocab items, which consist of a single word, features include a binary indicator for whether the word is real, its length, and binary indicators for its presence in CEFR-specific wordlists. We also consider log frequency, log frequency-rank, capitalization frequency, and n-gram features from COCA. For ViC items, we use surface features such as the number of missing characters and vowel proportion in the missing part, along with log frequencies from COCA and its sub-corpora. We include RoBERTa embeddings of the masked passage (reduced to 10 dimensions using PCA), sentence-level log frequencies, the normalized position of the damaged word, and the conditional probability of the correct word based on visible letters and word length.  The use of foundation model features is indispensable to our calibration performance.

{\bf Administration of Vocabulary Items.}
The control condition in our experiments is the previous version of the DET practice test, which includes Y/N Vocab items but no ViC items.
The control Y/N vocab items are arranged in a card format consisting of 18 words per card (roughly half real and half fake).
The test taker is instructed to identify the real words within the card, and thoughout the test there are 4 such cards, for a total of 72 Y/N Vocab items.
The card format has the disadvantage that it doesn't enable strong adaptivity, since the words displayed on the card are predetermined.

The administration algorithm for treatment conditions is the following.
18 Y/N Vocab items and 9 ViC items are administered in sequence at the beginning of DET test sessions.
For each Y/N Vocab administration event, we draw a Bernoulli$(1/2)$ to determine if the word displayed is real or fake.
Then the real or fake item is selected using BanditCAT V1 from among the eligible items.
Only items that have not yet been administered are eligible.
The session continues with 9 ViC items with 20 second time limits after an instructional screen. 

\section{Experimental Results}

\subsection{DET Experiments}

Table \ref{tab:experiments} summarizes a series of experiments conducted for the DET practice tests conducted between 2023-01-23 and 2023-02-19. 
The primary purpose of these experiments are to converge on a final candidate for DET V8.
While these experiments were important steps along the way to obtaining this launch candidate, none of the treatment conditions listed here are operational on the DET.
None of these experiments or their metrics are reflective of the certified DET.
All treatment conditions listed here use AutoIRT + BanditCAT V1 as explained above.
The experiments are grouped by experimental blocks (E1 to E5) and contain control and treatment conditions (e.g., T1, T2, T3, etc.). 
The control conditions (labeled as "C") across different experimental blocks use 4x Y/N (yes/no) vocab items, while the treatment conditions only vary in how they administer other aspects of the test such as other item types
Some experiments, such as those marked with "*", denote adjusted calibrations of the AutoIRT method, with changes to the model or the downstream administration from a prior condition. 
Experiment E5 includes ViC items in addition to the Y/N Vocab items because this is the first experiment that used AutoIRT + BanditCAT V1 for ViC items.
In these treatment conditions, the ViC administration was not dependent on the responses to the Y/N Vocab items.
Each condition has a different number of participants (N) ranging from roughly 4,000 to over 14,000, indicating robust data collection efforts across different time spans in early 2023.
In all of these cases, the items were calibrated in the jump-start setting, with a limited amount of response data for both of these item types.
This preliminary response data was gathered from practice test sessions that were administered a treatment condition in a prior experiments (that we will not be discussing here).
In all experiments, we set the exposure control parameter using the aforementioned nearest neighbor matching to hit a reasonable target.

\begin{table}[ht]
\centering
\begin{tabular}{|l|l|p{1cm}|p{1cm}|l|p{7cm}|}
\hline
\textbf{Exper.} & \textbf{Cond.} & \textbf{Min Date} & \textbf{Max Date} & \textbf{N} & \textbf{Description}\\ \hline
E1 & C   & 01-23 & 01-26 & 6304 & Control, 4x Y/N Vocab Items\\ \hline
E1 & T1  & 01-23 & 01-26 & 6535 & Y/N Vocab uses AutoIRT and BanditCAT V1\\ \hline
E1 & T2  & 01-23 & 01-26 & 6050 & Y/N Vocab uses AutoIRT and BanditCAT V1, different downstream admin. from T1\\ \hline
E2 & C   & 01-26 & 01-31 & 10560 & Control, 4x Y/N Vocab Items, no ViC \\ \hline
E2 & T1* & 01-26 & 01-31 & 10311 & Adjusted calibration of T1\\ \hline
E2 & T2* & 01-26 & 01-31 & 10292 & Adjusted calibration of T2\\ \hline
E3 & C   & 01-31 & 02-05 & 14086 & Control, 4x Y/N Vocab Items, no ViC\\ \hline
E3 & T3  & 01-31 & 02-05 & 13207 & Same as T2* above with different downstream admin.\\ \hline
E3 & T2* & 01-31 & 02-05 & 13855 & Same as T2* above\\ \hline
E4 & C   & 02-12 & 02-16 & 7498 & Control, 4x Y/N Vocab Items, no ViC\\ \hline
E4 & T4  & 02-12 & 02-16 & 7021 & Same as T2* above with different downstream admin.\\ \hline
E4 & T2* & 02-12 & 02-16 & 14118 & Same as T2* above\\ \hline
E5 & C   & 02-16 & 02-19 & 4834 & Control, 4x Y/N Vocab Items, no ViC\\ \hline
E5 & T5  & 02-16 & 02-19 & 4301 & Y/N Vocab and ViC uses AutoIRT and BanditCAT V1\\ \hline
E5 & T6  & 02-16 & 02-19 & 8638 & Y/N Vocab and ViC uses AutoIRT and BanditCAT V1, different downstream admin. from T5\\ \hline
\end{tabular}
\caption{A description of the DET practice test experiments and their conditions.}
\label{tab:experiments}
\end{table}

\subsection{Evaluation Metrics}

Retest reliability (RR) is a common performance measure, defined as the Pearson correlation between scores when the same user takes the test twice. 
In classical test theory, the observed score is seen as the true score plus noise. The standard error of measurement (SEM), which estimates how much a score deviates from the true score, is related to RR by the formula:
\[
S_E = S_X \sqrt{1 - RR},
\]
where $S_X$ is the population score standard deviation, and $S_E$ is the error standard deviation. 
For example, increasing RR from 0.5 to 0.6 reduces SEM by 11.8\%.
In practice tests, the reliability is typically smaller than in certified tests where the test taker's motivation is higher.
Another common metric for a item type score (such as the Y/N Vocab score) is the correlation between that item types score and the overall score for the test.
The DET consists of 14 task types ranging from interactive listening and reading to dictation \citep{cardwell2022duolingo}.
We also consider the maximum exposure rates, that is over all of the test sessions, what is the maximum frequency of an item being selected ($f_i$) as a ratio of the total number of administration events for items of that type.

\subsection{Results Discussion}

\begin{table}[ht]
\centering
\begin{tabular}{|c|c|c|c|c|c|}
\hline
\textbf{Experiment} & \textbf{Cond.} & \textbf{Task} & \textbf{RR} & \textbf{Score Corr.} & \textbf{Max exposure} \\ \hline
E1 & C   & Y/N Vocab       & 0.689 & 0.813 & 1.00\% \\ \hline
E1 & T1  & Y/N Vocab       & 0.535 & 0.723 & 0.56\% \\ \hline
E1 & T2  & Y/N Vocab       & 0.550 & 0.733 & 0.57\% \\ \hline
E2 & C   & Y/N Vocab       & \textbf{0.724} & 0.824 & 0.98\% \\ \hline
E2 & T1* & Y/N Vocab       & 0.579 & 0.752 & 0.65\% \\ \hline
E2 & T2* & Y/N Vocab       & \textbf{0.587} & 0.753 & 0.67\% \\ \hline
E3 & C   & Y/N Vocab       & 0.722 & \textbf{0.825} & 0.95\% \\ \hline
E3 & T3  & Y/N Vocab       & 0.571 & 0.743 & 0.64\% \\ \hline
E3 & T2* & Y/N Vocab       & 0.567 & 0.758 & 0.61\% \\ \hline
E4 & C   & Y/N Vocab       & 0.713 & 0.821 & 0.91\% \\ \hline
E4 & T4  & Y/N Vocab       & 0.538 & 0.748 & 0.74\% \\ \hline
E4 & T2* & Y/N Vocab       & 0.543 & 0.741 & 0.74\% \\ \hline
E5 & C   & Y/N Vocab       & 0.713 & 0.802 & 0.97\% \\ \hline
E5 & T5  & Y/N Vocab       & 0.535 & 0.738 & 0.66\% \\ \hline
E5 & T6  & Y/N Vocab       & 0.552 & \textbf{0.759} & 0.64\% \\ \hline
E5 & T5  & ViC             & 0.659 & 0.737 & 1.10\% \\ \hline
E5 & T6  & ViC             & \textbf{0.661} & \textbf{0.742} & 1.01\% \\ \hline
\end{tabular}
\caption{Reliability, validity, and exposure results of DET practice test experiments. We have bolded the maximum metrics for Y/N Vocab, ViC tasks, and control (for Y/N Vocab). All metrics listed are not reflective of the DET certified test or the current version of the practice test.}
\label{tab:metrics}
\end{table}

Table \ref{tab:metrics} gives the results of our experiments in the DET practice test.
When it comes to the Y/N Vocab and ViC sections, the treatment conditions do not differ substantially with the exception of going from $T1, T2$ to $T1^*, T2^*$.
In that case, the recalibration improved scoring and administration thus increasing the reliability metrics.
In general, the control condition has significantly higher reliabilities and score correlations, but this is due to the fact that it has 4x the number of items as the treatment.
We see that the loss in reliability for Y/N vocab when we reduce the number of items from 72 to 18 does not result in a similar loss of reliability (which we would expect to roughly half).
Similarly, the drop in score correlations is not as significant as we might expect given the dramatic reduction in the number of items.
We see that we were very effectively able to control exposure, since we did not see a significant change from the control condition.

While there is no control comparison for the ViC item type, we see that with 9 items we get a reliability of 0.66, which exceeds that for the 18 Y/N vocab items.
It should be noted that the combination of these two item type scores will get significantly higher reliability ($0.8$ for E5-T6 for the average scores).
The original time allotted to vocabulary items in the control condition is 4 minutes.
The time allotted for Y/N Vocab and ViC combined is 4:30, yet there is a very substantial total reliability increase.

\section{Conclusions}

This work demonstrates a general approach to calibrating and administering an adaptive test with minimal responses and limited modeling effort.
It outlines the AutoIRT method for obtaining item parameter estimates from complex item features, as well as foundation model embeddings.
Specifically, we use the RoBERTa embeddings for items that contain passages (vocabulary-in-context).
We also introduce the BanditCAT framework, which casts computerized adaptive tests in the contextual bandit setting.
We provide an algorithm that is a first attempt at using this framework, BanditCAT V1 --- a Thompson sampling approach to test administration.

While this paper does introduce the BanditCAT framework, BanditCAT V1 is just the first algorithm that utilizes this approach.
However, this algorithm does not address a few major issues that plague adaptive testing.
First, it is not a complete Thompson sampling algorithm because it does not also sample the item parameters from a posterior distribution.
Because this is another critical source of uncertainty in the reward model (Fisher information), this would improve the balance between exploration and exploitation.
Second, the method for exposure control is specific to the 2PL and 3PL IRT models, a general purpose exposure control would extend this approach to other IRT models.
Third, this model only supports unidimensional $\theta$, and to extend this problem to multidimensional $\theta$ would require additional modifications.
This work is a precursor to a more complete treatment that addresses these issues.

\bibliography{banditcat}


\end{document}